\documentclass{amsart}
\usepackage[dvips]{graphicx}
\usepackage{amsmath}
\usepackage{amsfonts}
\usepackage{amssymb}
\usepackage{amscd}
\usepackage{t1enc}
\usepackage{natbib}
\usepackage{url}
\usepackage[latin2]{inputenc}
\newtheorem{thm}{Theorem}[section]

\newtheorem{defn}[thm]{Definition}

\renewcommand{\b}[1]{\mathbf{#1}}
\newcommand{\Real}{\mathbb R}

\bibpunct{(}{)}{;}{a}{,}{,}

\begin{document}

\title[Learning Pac-Man and the LCMP framework]{Low-Complexity Modular Policies: Learning to Play Pac-Man and a New Framework beyond MDPs}

\author{Istv{\'a}n Szita and Andr{\'a}s L{\H{o}}rincz }

\begin{abstract}
In this paper we propose a method that learns to play Pac-Man. We
define a set of high-level observation and action modules. Actions
are temporally extended, and multiple action modules may be in
effect concurrently. A decision of the agent is represented as a
rule-based policy. For learning, we apply the cross-entropy
method, a recent global optimization algorithm. The learned
policies reached better score than the hand-crafted policy, and
neared the score of average human players. We argue that learning
is successful mainly because (i) the policy space includes the
combination of individual actions and thus it is sufficiently
rich, (ii) the search is biased towards low-complexity policies
and low complexity solutions can be found quickly if they exist.
Based on these principles, we formulate a new theoretical
framework, which can be found in the Appendix as supporting
material.
\end{abstract}

\maketitle

\section{Introduction} \label{s:intro}

During the last two decades, reinforcement learning has reached a
mature state, and has been laid on solid foundations. We have a
large variety of algorithms, including value-function based,
direct policy search and hybrid methods
\citep{Sutton98Reinforcement,Bertsekas96Neuro-Dynamic}. The basic
properties of many such algorithms are relatively well understood
(e.g. conditions for convergence, complexity, effect of various
parameters etc.), although it is needless to say that there are
still lots of important open questions. There are also plenty of
test problems (like various maze-navigation tasks, pole-balancing,
car on the hill etc.) on which the capabilities of RL algorithms
have been demonstrated, and the number of successful large-scale
RL applications is also growing steadily. However, there is still
a sore need for more successful applications to validate the place
of RL as a major branch of artificial intelligence.

We think that games (including the diverse set of classical board
games, card games, modern computer games etc.) are ideal test
environments for reinforcement learning. Games are intended to be
interesting and challenging for human intelligence and therefore,
they are ideal means to explore what artificial intelligence is
still missing. Furthermore, most games fit well into the RL
paradigm: they are goal-oriented sequential decision problems,
where each decision can have long-term effect. In many cases,
hidden information, random events, unknown environment, known, or
unknown players account for (part of) the difficulty of playing
the game. Such circumstances are in the focus of the reinforcement
learning idea. They are also attractive for testing new methods:
the decision space is huge in most cases, so finding a good
strategy is a challenging task.

There is another great advantage of games as test problems: the
rules of the games are fixed, so the danger of `tailoring the task
to the algorithm' -- i.e., to tweak the rules and/or the
environment so that they meet the capabilities of the proposed RL
algorithm -- is reduced, compared, e.g., to various maze
navigation tasks.

RL has been tried in many classical games, including checkers
\citep{Samuel59Some}, backgammon \citep{Tesauro94TDGammon}, and
chess \cite{Baxter01Reinforcement}. On the other hand, modern
computer games got into the spotlight only recently, and there are
not very many successful attempts to learn them with AI tools.
Notable exceptions are, e.g.,  role-playing game \emph{Baldur's
Gate} \citep{Spronck03Online}, real-time strategy game
\emph{Wargus} \citep{Ponsen04Improving}), and possibly,
\emph{Tetris} \citep{szita06learning}. These games are also
interesting from the point of view of RL, as they catch different
aspects of human intelligence: instead of deep and wide logical
deduction chains, most modern computer games need short-term
strategies, but many observations have to be considered in
parallel, and both the observation space and the action space can
be huge.

In this spirit, we decided to investigate the arcade game Pac-Man.
The game is interesting on its own, as it is largely unsolved, but
also imposes several important questions in RL, which we will
overview in Section~\ref{s:discussion}. We will show that a hybrid
approach is more successful than either tabula rasa learning or a
hand-coded strategy alone. We will provide hand-coded high-level
actions and observations, and the task of RL is to learn how to
combine them into a good policy. We will apply rule-based policies
because they are easy to interpret, and it is easy to include
human domain-knowledge. For learning, we will apply the
cross-entropy method, a recently developed general optimization
algorithm.

In the next section we overview the Pac-Man game and the related
literature. We also investigate the emerging questions upon
casting this game as a reinforcement learning task. In sections
\ref{s:rulebased} and \ref{s:ce} we give a short description of
rule-based policies and the cross-entropy optimization method,
respectively. In section \ref{s:experiments} we describe the
details of the learning experiments, and in section
\ref{s:results} we present our results. Finally, in section
\ref{s:discussion} we summarize and discuss our approach with an
emphasis on its implications for other RL problems.

\section{Pac-Man and reinforcement learning}
\label{s:pacman_and_rl}

\subsection{The Pac-Man game}

\begin{figure}
\centering
\includegraphics[width=6cm,height=8cm]{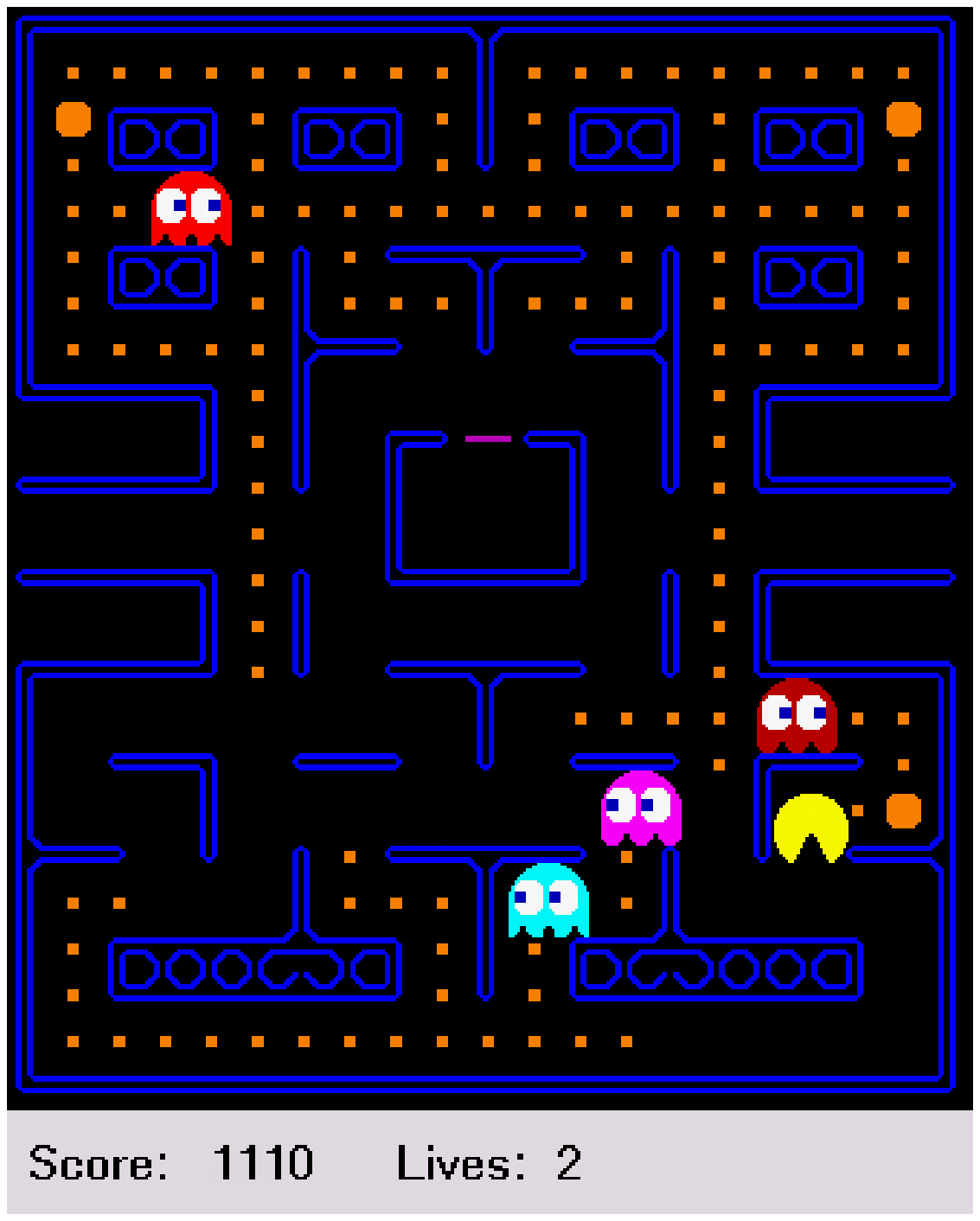}
 \caption{\textbf{Pac-Man.}}\label{fig:pacman_board}
\end{figure}

The video-game Pac-Man was first released in 1979, and reached
immense success, it is considered to be one of the most popular
video games to date \citep{Wiki:Pac-Man}.

The player maneuvers Pac-Man in a maze (see
Fig.~\ref{fig:pacman_board}), while `eating' the dots in the maze.
There are 174 dots, each one is worth 10 points. A level is
finished when all the dots are eaten. To make things more
difficult, there are also four ghosts in the maze `who' try to
catch Pac-Man, and if they succeed, Pac-Man loses a `life'.
Initially, `he' has three lives, and gets an extra life after
reaching 10,000 points.

There are four power-up items in the corners of the maze, called
\emph{power dots} (worth 40 points). After Pac-Man eats a power
dot, the ghosts turn blue for a short period, they slow down and
try to escape from Pac-Man. During this time, Pac-Man is able to
eat them, which is worth 200, 400, 800 and 1600 points,
consecutively. The point values are reset to 200 each time another
power dot is eaten, so it is advantageous for the player to eat
all four ghosts per power dot. After being eaten, ghosts are
`reborn' in the center of the maze.

Our investigations are restricted to learning an optimal policy
for the first level, so the maximum achievable score is
$174\cdot10 + 4\cdot40 + 4\cdot(200+400+800+1600) =
13,900$.\footnote{The rules of the original Pac-Man game are
slightly different. The above description applies to the
open-source Pac-Man implementation of \citet{Courtillat01Pacman}.
The two versions are about equivalent in terms of complexity and
entertainment value.}

In the original version of Pac-Man, ghosts move on a complex but
deterministic route, so it is possible to learn a deterministic
action sequence that does not require any observations. Many such
\emph{patterns} were found by enthusiastic players. In our
implementation, ghosts moved randomly in 20\% of the time and
straight towards Pac-Man in the remaining 80\%, but ghosts may not
turn back (in accordance with the original implementation). This
way, there is no single optimal action sequence, observations are
required for optimal decision making. Similar methods of
randomization are implemented in many Pac-Man's sequels (e.g.,
Ms.~Pac-Man).

\subsection{Previous work on Pac-Man}



Although the game can be properly formalized as a finite MDP, the
resulting model would have about $10^{70}$ states. The learning
task is hard even with approximation techniques, so the only RL
approach known to us \citep{bonet99learning} restricts
observations to a $10\times 10$ window centered at Pac-Man.
Through a series of increasingly difficult learning tasks, they
were able to teach basic pellet-collecting and ghost-avoidance
behaviors in greatly simplified versions of the game: they used
simple mazes containing no power pellet and only one ghost.

There have been several other attempts using genetic algorithms,
and the only full-scale Pac-Man learner that we know uses genetic
algorithms with hand-crafted features and it applies a neural
network position evaluator \citep{lucas05evolving}.

\subsection{Pac-Man as an RL task}

Pac-Man meets all the criteria of a reinforcement learning task.
The agent has to make a sequence of decisions that depend on its
observations. The environment is stochastic (the path of ghosts is
unpredictable). There is also a well-defined reward function (the
score for eating things), and actions influence the collected
reward in the remote future.

The full description of the state would include (1) whether the
dots have been eaten (one bit for each dot and one for each power
dot), (2) the position and direction of Pac-Man, (3) the position
and direction of the four ghosts, (4) whether the ghosts are blue
(one bit for each ghost), (5) the number of lives left. The
resulting state space is enormous, so some kind of function
approximation or feature-extraction are necessary for RL.

The action space seems less problematic, as there are only four
basic actions: go north/south/east/west. However, a typical game
consists of multiple hundreds of steps, so the number of possible
combinations is still enormous. This indicates the need for
temporally extended actions.

We have a moderate amount of domain knowledge about Pac-Man: on
one hand, it is quite easy to define high-level observations and
action modules that are potentially useful. On the other hand,
constructing a well-performing policy seems much more difficult.
Therefore, we chose a hybrid approach: we use domain knowledge to
preprocess the state information and to define action modules, and
combine them into a rule-based policy. However, we use policy
search reinforcement learning to learn the proper combination.

\section{Rule-based policies} \label{s:rulebased}

In a basic formulation, a rule is a sentence of the form
"\texttt{if \emph{Condition} holds, then do \emph{Action}}". A
rule-based policy is a set of rules with some mechanism for
breaking ties, i.e., to decide which rule is executed, if there
are multiple rules with satisfied conditions.

Rule-based policies are human-readable, it is easy to include
domain knowledge, and they are able to represent complex
behaviors. For these reasons, they are often used in many areas of
artificial intelligence, e.g. \citep{Spronck03Online}.

In order to apply rule-based policies to Pac-Man, we need to
specify four things: (1) what are the possible actions (2) what
are the possible conditions and how are they constructed from
observations, (3) How to make rules form consitions and actions,
and (4) how to combine the rules into policies. These will be
described in the following sections.

\subsection{Action modules}

\begin{table}[h]
\begin{center}
\caption{List of action modules used for rule construction.}
 \label{t:actions} \vskip 0.15in
\begin{tabular}{l|p{80mm}}
   Name  & Description \\[2pt]
\hline \hline \rule{0pt}{12pt}%
  \texttt{ToDot} & Go towards the nearest dot. \\
  \texttt{ToPowerDot} & Go towards the nearest power dot. \\
  \texttt{FromPowerDot} & Go in direction opposite to the nearest power dot. \\
  \texttt{ToEdGhost} & Go towards the nearest edible (blue) ghost. \\
  \texttt{FromGhost} & Go in direction opposite to the nearest ghost. \\
  \texttt{ToSafeJunction} & For all four directions, the "safety" of
  the nearest junction is estimated in that direction. If Pac-Man is $n$ steps away
  from the junction and the nearest ghost is $k$ steps away, then the safety
  value of this junction is $n-k$. A negative value means that Pac-Man
  possibly cannot reach that junction. Pac-Man goes towards the maximally safe junction. \\
  \texttt{FromGhostCenter} & Go in a direction which maximizes the Euclidean
  distance from the geometrical center of ghosts. \\
  \texttt{KeepDirection} & Go further in the current direction (or turn
  right/left if that is impossible).\\
  \texttt{ToLowerGhostDensity} & Each ghost defines a density cloud
  (with radius = 10 and linear decay). Pac-Man goes in the direction
  where the cumulative ghost density decreases fastest.\\
  \texttt{ToGhostFreeArea} & Chooses a location on the board where the minimum
  ghost distance is largest, and heads towards it on the shortest path.\\
 \hline
\end{tabular}
\end{center}
\end{table}

We can define a list of potentially useful action modules for
Pac-Man (see Table~\ref{t:actions}). Some of these are intuitive,
while the last five were deduced by playing and analyzing the
game.

Note that these modules are not exclusive. For example, while
escaping from the ghosts, Pac-Man may prefer the route where more
dots can be eaten, or it may want to head towards a power dot.
Without the possibility of such parallel actions, the performance
of the Pac-Man agent may be reduced, and preliminary experiments
showed that this is the case, indeed.

We need a mechanism for conflict resolution, because different
action modules may suggest different directions. We do this by
assigning \emph{priorities} to the modules. When the agent
switches on an action module, he also decides its priority. This
is also a decision, and learning this decision is part of the
learning task.

We implemented this with the following mechanism: a decision of
the agent concerns action modules: the agent can either
\emph{switch on} or, \emph{switch off} an action module. That is,
the agent is able to use any subset of the action modules -- at
least in principle --, instead of selecting a single one at each
time step. Basically, the module(s) with highest priority
decide(s) the direction of Pac-Man. If there are more than one
equally ranked directions, or modules with equal priority suggest
different directions, then lower-priority modules are checked. If
the direction cannot be decided after checking all switched-on
modules, then a random direction is chosen.

\subsection{Conditions and Observations}

\begin{table}[h]
\begin{center}
\caption{List of observations used for rule construction.}
 \label{t:observations} \vskip 0.15in
\begin{tabular}{l|p{80mm}}
   Name  & Description \\[2pt]
\hline \hline \rule{0pt}{12pt}%
  \texttt{Constant} & Constant 1 value. \\
  \texttt{NearestDot} & Distance of nearest dot. \\
  \texttt{NearestPowerDot} & Distance of nearest power dot. \\
  \texttt{NearestGhost} & Distance of nearest ghost. \\
  \texttt{NearestEdGhost} & Distance of nearest edible (blue) ghost. \\
  \texttt{MaxJunctionSafety} & For all four directions, the "safety" of
  the nearest junction in that direction is estimated, as defined in the description
  of action "ToSafeJunction". The observation returns the value
  of the maximally safe junction. \\
  \texttt{GhostCenterDist} & Euclidean distance from the geometrical center
  of ghosts. \\
  \texttt{DotCenterDist} & Euclidean distance from the geometrical center
  of uneaten dots.\\
  \texttt{GhostDensity} & Each ghost defines a density cloud
  (with radius = 10 and linear decay). Returns the value of the
  cumulative ghost density.\\
 \hline
\end{tabular}
\end{center}
\end{table}

Similarly to actions, we can easily define a list of observations
which are potentially useful for decision making. The observations
and their descriptions are summarized in
Table~\ref{t:observations}. Distances denote the "length of the
shortest path", unless noted otherwise. Distance to a particular
object type is `infinite' if no such object exists at that moment.

Now we have the necessary tools for defining the conditions of a
rule. A typical condition is true if its observations are in a
given range. We note that the status of each action module is also
important for proper decision making. For example, the agent may
decide that if a ghost is very close, then it switches off all
modules except the escape module. Therefore we allow conditions
that check whether an action module is `on' or `off'.

For the sake of simplicity, conditions were restricted to have the
form \mbox{"\texttt{[observation] > [value]}"},
\mbox{"\texttt{[observation] < [value]}"},
\mbox{"\texttt{[action]+}"}, \mbox{"\texttt{[action]-}"}, or the
conjunction of such terms. For example,
$$\mbox{"\texttt{(NearestDot<5) and (NearestGhost>8) and
(FromGhost+)}"}$$ is a valid condition for our rules.

\subsection{Constructing rules from conditions and actions}

Now, we have conditions and actions. A rule has the form:
"\texttt{if [Condition] holds, then do [Action]}". For
example,\smallskip\smallskip\smallskip
\newline "\mbox{\texttt{if (NearestDot<5) and (NearestGhost>8) and (FromGhost+)}}\\
 \mbox{\texttt{\hspace{80mm}then FromGhostCenter+}}"

 \noindent is a valid
 rule. In all of our experiments, we considered only rules with at
 most three conditions.

\subsection{Constructing policies from rules}

Decision lists are standard forms of constructing policies from
single rules. This is the approach we pursue here, too. Decision
lists are simply lists of rules, together with a mechanism that
decides the order in which the rules are checked.

We assign priorities to each rule. When the agent has to make a
decision, it checks its list of rules, starting with the highest
priority ones. If the conditions of a rule are fulfilled, then the
corresponding action is executed, and the decision-making process
halts.

Note that in principle, the priority of a rule can be different
from the priority of action modules. However, for the sake of
simplicity, we make no distinction: if a rule with priority $k$
\emph{switches on} an action module, then the priority of the
action module is also taken as $k$. Intuitively, this makes sense:
if an important rule is activated, then its effect should also be
important. Naturally, if a rule with priority $k$ \emph{switches
off} a module, then it is executed, regardless of the priority of
the module.

\subsection{An example}

Let us consider the example shown in Table~\ref{t:policyexample}.
This is a rule-based policy for the Pac-Man agent.
\begin{table}[h]
\begin{center}
\caption{\textbf{A sample rule-based policy.} Bracketed numbers
denote priorities, \texttt{[1]} is the highest priority.}
 \label{t:policyexample} \vskip 0.15in
\begin{tabular}{l|c|p{80mm}}
   Rule No.  & Priority &  Rule \\[2pt]
\hline \hline \rule{0pt}{12pt}%
    \texttt{Rule 1} & \texttt{[1]} & \texttt{if  (NearestGhost<4) then FromGhost+} \\
    \texttt{Rule 2} & \texttt{[1]} & \texttt{if  (NearestGhost>7) and (JunctionSafety>4) then FromGhost-} \\
    \texttt{Rule 3} & \texttt{[2]} & \texttt{if  (NearestEdGhost>99) then ToEdGhost-} \\
    \texttt{Rule 4} & \texttt{[2]} & \texttt{if  (NearestEdGhost<99) then ToEdGhost+} \\
    \texttt{Rule 5} & \texttt{[3]} & \texttt{if  (Constant>0) then KeepDirection+} \\
    \texttt{Rule 6} & \texttt{[3]} & \texttt{if  (FromPowerDot-) then ToPowerDot+} \\
    \texttt{Rule 7} & \texttt{[3]} & \texttt{if  (GhostDensity<1.5) and (NearestPowerDot<5) then FromPowerDot+} \\
    \texttt{Rule 8} & \texttt{[3]} & \texttt{if  (NearestEdGhost>99) then FromPowerDot-} \\
    \texttt{Rule 9} & \texttt{[3]} & \texttt{if  (NearestPowerDot>10) then FromPowerDot-} \\
 \hline
\end{tabular}
\end{center}
\end{table}

The first two rules manage ghost avoidance: if a ghost is too
close, then the agent should flee, and should do so until it gets
to a safe distance. Ghost avoidance has priority over any other
activities. The next two rules regulate that if there is an edible
ghost on the board, then the agent should chase it (the value of
\texttt{NearestEdGhost} is infinity ($>99$) if there are no edible
ghosts, but it is $\leq 41$ on our board, if there are). This
activity has also relatively high priority, because eating ghosts
is worth lots of points, but it must be done before the blue color
of the ghost disappears, so it must be done quickly. The fifth
rule says that the agent should not turn back, if all directions
are equally good. This rule prevents unnecessary zigzagging (when
no dots are eaten), and it is surprisingly effective. The
remaining rules tweak the management of power dots. Basically, the
agent prefers to eat a power dot. However, if there are blue
ghosts on the board, then a power dot resets the score counter to
200, so it is a bad move. Furthermore, if ghost density is low
around the agent, then most probably it will be hard to collect
all of the ghosts, so it is preferable to wait with eating the
power dot.

\begin{figure}[h!]
\centering
\includegraphics[width=12cm]{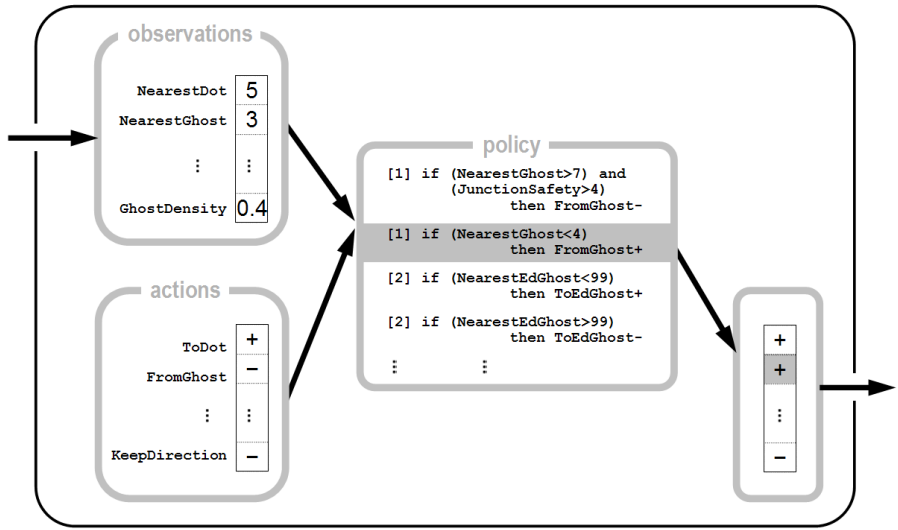}
 \caption{\textbf{Decision-making Mechanism of Pac-Man agent.}}\label{fig:decisionmaking}
\end{figure}

The mechanism of decision making is depicted in
Fig~\ref{fig:decisionmaking}. In short, the (hidden) state-space
is the world of the Pac-Man and the Ghosts. The dynamics of this
(hidden) state-space determines the vector of observations, which
can be checked by the conditions. If the conditions of a rule are
satisfied, the corresponding action module is switched on or off.
As a consequence, multiple actions may be in effect at once. For
example, the decision depicted in Fig.~\ref{fig:decisionmaking}
sets two actions to work together.

\subsection{Learning rule-based policies by policy search}

We will perform policy search RL in the space of rule-based
policies. Our algorithm will construct policies according to its
parameter set. The policies will be tested in the environment, by
using them to control Pac-Man and measure the collected rewards.
The results of these tests are then used to improve the parameter
set, and consequently, the policy construction procedure.

\section{The cross-entropy method} \label{s:ce}

Our goal is to learn a rule-based policy that has the form
described in the previous section, by performing policy search in
the space of all legal rule-based policies. For this search we
apply the \emph{cross-entropy method}, a recently published global
optimization algorithm \citep{rubinstein99crossentropy}. Below we
summarize the mechanism of this method briefly.

\subsection{The general form of the algorithm}

The cross-entropy (CE) method is a general algorithm for
(approximately) solving global optimization tasks of the form
\begin{equation}
  \b x^* := \arg\max_{\b x} f(\b x).
\end{equation}
where $f$ is a general objective function (e.g., we do not need to
assume continuity or differentiability). While most optimization
algorithms maintain a single candidate solution $x(t)$ in each
time step, CE maintains a \emph{distribution} over possible
solutions. From this distribution, solution candidates are drawn
at random. This is essentially random guessing, but with a nice
trick it is turned into a highly effective optimization method.

Random guessing is an overly simple `optimization' method: we draw
many samples from a fixed distribution $g$, then select the best
sample as an estimation of the optimum. In the limit case of
infinitely many samples, random guessing finds the global optimum.
We have two notes here: (i) as it has been shown by
\cite{wolpert97nofree}, for the most general problems, uniform
random guessing is not worse than any other method, (ii)
nonetheless, for practical problems, uniform random guessing can
be extremely inefficient. Thus, random guessing is safe to start
with, but as one proceeds with the collection of experiences, it
should be limited as much as possible.

The efficiency of random guessing depends greatly on the
distribution $g$ from which the samples are drawn. For example, if
$g$ is sharply peaked around $\b x \neq \b x^*$, then a tremendous
number of examples are needed to get a good estimate of the global
optimum. The case is the opposite, if the distribution is sharply
peaked at $\b x^*$: very few samples may be sufficient to get a
good estimate. Naturally, finding a good distribution is at least
as hard as finding $\b x^*$.

The idea of CE is that after drawing moderately many samples from
distribution $g$, we may not be able to give an acceptable
approximation of $\b x^*$, but we may still obtain a \emph{better
sampling distribution}. We will pick $g$ from a family of
parameterized distributions, denoted by $\mathcal G$, and describe
an algorithm that iteratively improves the parameters of the
distribution $g$.

For each $\gamma\in\Real$, the set of high-valued samples,
\[
  \hat L_\gamma := \{ \b x^{(i)} \mid f(\b x^{(i)}) \geq \gamma, 1\leq i\leq N \},
\]
provides an approximation to the level set
\[
  L_\gamma := \{ \b x \mid f(\b x) \geq \gamma \}.
\]
Let $U_\gamma$ be the uniform distribution over the level set
$L_\gamma$. For large values of $\gamma$, this distribution will
be peaked around $\b x^*$, so it would be suitable for random
sampling. There are two problems with that: (i) for large $\gamma$
values $\hat L_\gamma$ will contain very few points (possibly
none), making accurate approximation impossible, and (ii) the
level set $L_\gamma$ is usually not a member of the parameterized
distribution family.

CE avoids the first problem by making a compromise in the choice
of $\gamma$: it prefers large improvements, so does not set
$\gamma$ too low, but it does not set $\gamma$ too high either in
order to keep plenty of samples in $\hat L_\gamma$. This
compromise is achieved as follows: CE chooses a ratio
$\rho\in[0,1]$ and adjusts $\hat L$ to be the set of the best
$\rho \cdot N$ samples. This corresponds to setting $\gamma :=
f(\b x^{(\rho\cdot N)})$, provided that the samples are arranged
in decreasing order of their values. The best $\rho \cdot N$
samples are called the \emph{elite samples}. In practice, $\rho$
is typically chosen from the range $[0.02, 0.1]$.

The other problem is solved by changing the goal of the
approximation: CE chooses the distribution $g$ \emph{from the
distribution family $\mathcal{G}$} that approximates best the
empirical distribution over $\hat L_\gamma$. The best $g$ is found
by minimizing the distance of $\mathcal{G}$ and the uniform
distribution over the elite samples. The measure of distance is
the \emph{cross-entropy distance} (often called Kullback-Leibler
divergence). The cross-entropy distance of two distributions $g$
and $h$ is defined as
\begin{equation}
  D_{CE}(g||h) = \int g(\b x) \log \frac{g(\b x)}{h(\b x)} \textrm{ d}\b x
\end{equation}

The general form of the cross-entropy method is summarized in
Table~\ref{t:CE_method}. It is known that under mild regularity
conditions, the CE method converges with probability 1
\citep{margolin04onthe}. Furthermore, for a sufficiently large
population, the global optimum is found with high probability.

\begin{table}[h]
 \hrule \vskip1pt \hrule \vskip1mm
\begin{tabbing}
 xxx \= xxx \= xxxxxxxxxxx \= xxxxxxxxxxxxxxxxxxxxxx \= \kill
 input: $\mathcal{G}$                        \>\>\>\> \%parametrized distribution family\\
 input: $g_0 \in \mathcal{G}$                \>\>\>\> \%initial distribution\\
 input: $N$                                  \>\>\>\> \%population size\\
 input: $\rho$                               \>\>\>\> \%selection ratio\\
 input: $T$                                  \>\>\>\> \%number of iterations\\
 for $t$ from 0 to $T-1$,                    \>\>\>\> \%CE iteration main loop \\
 \> for $i$ from 1 to $N$,\\
 \>\> draw $\b x^{(i)}$ from distribution $g_t$      \>\> \%draw $N$ samples \\
 \>\> compute $f_i := f(\b x^{(i)})$                 \>\> \%evaluate them \\
 \> sort $f_i$-values in descending order \\
 \> $\gamma_{t+1} := f_{\rho\cdot N}$         \>\>\> \%level set
 threshold \\
 \> $E_{t+1}:=\{\b x^{(i)} \mid f(\b x^{(i)})\geq\gamma_{t+1} \}$  \>\>\> \%get elite
 samples \\
 \> $g_{t+1} := \arg\min_{g\in\mathcal{G}} D_{CE}(g ||
 \textit{Uniform}(E_{t+1}) )$               \>\>\> \%get nearest distrib. from $\mathcal{G}$ \\
 end loop
\end{tabbing}
 \hrule \vskip1pt \hrule \vskip2mm
 \caption{Pseudo-code of the general cross-entropy method} \label{t:CE_method}
\end{table}

\subsection{The cross-entropy method for Bernoulli distribution}

For many parameterized distribution families, the parameters of
the minimum cross-entropy member can be computed easily from
simple statistics of the elite samples. We provide the formulae
for Bernoulli distributions, as these are needed for our purposes.
The derivations and a list of other discrete and continuous
distributions that have simple update rules can be found in the
tutorial of \citet{Boer04Tutorial}.

Let the domain of optimization be $D = \{0,1\}^m$, and each
component be drawn from independent Bernoulli distributions, i.e.
$\mathcal{G} = \textit{Bernoulli}^m$. Each distribution $g\in
\mathcal G$ is parameterized with an $m$-dimensional vector $\b p
= (p_{1}, \ldots, p_{m})$. When using $g$ for sampling, component
$j$ of the sample $\b x \in D$ will be
\begin{equation}
x_j = \left\{%
\begin{array}{ll}
    1, & \hbox{with probability $p_j$;} \\
    0, & \hbox{with probability $1-p_j$.} \\
\end{array}%
\right.
\end{equation}

After drawing $N$ samples $\b x^{(1)}, \ldots \b x^{(N)}$ and
fixing a threshold value $\gamma$, let $E$ denote the set of elite
samples, i.e.,
\begin{equation}
    E := \{ \b x^{(i)} \mid  f(\b x^{(i)}) \geq \gamma \}
\end{equation}
With this notation, the distribution $g'$ with minimum CE-distance
from the uniform distribution over the elite set has the following
parameters:
\begin{eqnarray}
    \b p' &:=& (p'_1, \ldots, p'_m), \quad\textrm{where}\\
    p'_j &:=& \frac{\sum_{\b x^{(i)}\in E} \chi(x^{(i)}_j \!=\! 1)}{\sum_{\b x^{(i)}\in E} 1} = \frac{\sum_{\b x^{(i)}\in E} \chi(x^{(i)}_j \!=\! 1)}{\rho\cdot
    N} \label{e:CEupdate_bernoulli}
\end{eqnarray}
In other words, the parameters of $g'$ are simply the
componentwise empirical probabilities of 1's in the elite set. For
the derivation of this rule, see \cite{Boer04Tutorial}.

Changing the distribution parameters from $\b p$ to $\b p'$ can be
too coarse, so in some cases, applying a step-size parameter
$\alpha$ is preferable. The resulting algorithm is summarized in
Table~\ref{t:CE_method_bernoulli}.

\begin{table}[h]
 \hrule \vskip1pt \hrule \vskip1mm
\begin{tabbing}
xxx \= xxx \= xxx \= xxx \= xxxxxxxxxxxxxxxxxxxxxx \= \kill
 \> input: $\b p_0 = (p_{0,1},\ldots,p_{0,m})$                \>\>\>\> \%initial distribution parameters\\
 \> input: $N$                                  \>\>\>\> \%population size\\
 \> input: $\rho$                               \>\>\>\> \%selection ratio\\
 \> input: $T$                                  \>\>\>\> \%number of iterations\\
 \> for $t$ from 0 to $T-1$,                    \>\>\>\> \%CE iteration main loop \\
 \>\> for $i$ from 1 to $N$,\\
 \>\>\> draw $\b x^{(i)}$ from $\textit{Bernoulli}^m(\b p_t)$      \>\> \%draw $N$ samples \\
 \>\>\> compute $f_i := f(\b x^{(i)})$                 \>\> \%evaluate them \\
 \>\> sort $f_i$-values in descending order \\
 \>\> $\gamma_{t+1} := f_{\rho\cdot N}$         \>\>\> \%level set
 threshold \\
 \>\> $E_{t+1}:=\{\b x^{(i)} \mid f(x^{(i)})\geq\gamma_{t+1} \}$  \>\>\> \%get elite
 samples \\
 \>\> $p'_j := \bigl(\sum_{\b x^{(i)}\in E} \chi(x^{(i)}_{j} \!=\! 1)\bigr)/(\rho\cdot N)$\\
        \>\>\>\>\> \%get parameters of nearest distrib. \\
 \>\> $p_{t+1,j} := \alpha\cdot p'_j + (1-\alpha)\cdot p_{t,j}$ \>\>\> \%update with step-size $\alpha$ \\
 \> end loop
\end{tabbing}
 \hrule \vskip1pt \hrule \vskip2mm
 \caption{Pseudo-code of the cross-entropy method for Bernoulli distributions} \label{t:CE_method_bernoulli}
\end{table}

We will also need to optimize functions over $D =
\{1,2,\ldots,K\}^m$ with $K>2$. In the simplest case,
distributions over this domain can be parameterized by $m\cdot K$
parameters: $\b p = (p_{1,1},\ldots, p_{1,K}; \ldots;
p_{m,1},\ldots, p_{m,K})$ with $0\leq p_{j,k} \leq 1$ and
$\sum_{k=1}^{K} p_{j,k} = 1$ for each $j$ (this is a special case
of the multinomial distribution).

The update rule of the parameters is essentially the same as
eq.~\ref{e:CEupdate_bernoulli} for the Bernoulli case:
\begin{eqnarray} \label{e:CEupdate_multinomial}
    p'_{j,k} &:=& \frac{\sum_{\b x^{(i)}\in E} \chi(x^{(i)}_j \!=\! k)}{\sum_{\b x^{(i)}\in E} 1} = \frac{\sum_{\b x^{(i)}\in E} \chi(x^{(i)}_j \!=\! k)}{\rho\cdot
    N}.
\end{eqnarray}
Note that constraint $\sum_{k=1}^{K} p'_{j,k} = 1$ is satisfied
automatically for each $j$.

\section{Description of experiments} \label{s:experiments}

All of the learning experiments used CE, which means drawing a
population of policies from some distribution, evaluating them by
playing the game, and updating the distribution parameters.

\subsection{Learning a policy from a hand-coded rulebase}

In the first experiment, we constructed a rulebase by hand. It
consisted of $K=40$ rules that were considered potentially useful.
The agent had to learn which rules to use, together with the
corresponding priorities.

From the rulebase, policies were constructed via the following
mechanism: a policy had $m=30$ \emph{rule slots}. For each $1\leq
i \leq m$, slot $i$ was filled with a rule from the rulebase with
probability $p_i$, and left empty with probability $1-p_i$. Each
slot had a fixed priority from the set $\{1,2,3\}$. For each
element of this set, we had 10 slots.\footnote{According to our
preliminary experiments, the quality of the learned policy did not
improve by increasing the priority set or the number of the
slots.} If it was decided that a slot should be filled, then a
particular rule $j$ ($1\leq j \leq K$) was selected with
probability $q_{i,j}$, where $\sum_{j=1}^K q_{i,j} = 1$ for each
slot $i\in \{1, \ldots, m\}$. As a result, policies could contain
up to 30 rules, but possibly much less.

Both the $p_i$ values and the $q_{i,j}$ values were learnt
simultaneously with the CE method
(Table~\ref{t:CE_method_bernoulli}), using the update rules
(\ref{e:CEupdate_bernoulli}) and (\ref{e:CEupdate_multinomial}),
respectively. This gave a total of $m + m\cdot K$ parameters to
optimize (although the effective number of parameters is much
less, because the $q_{i,j}$ values of unused slots are
irrelevant). Initial probabilities were set to $p_i = 1/2$ and
$q_{i,j} = 1/K$.

In each iteration, a population of $N=300$ policies were drawn
according to the actual probabilities. The value of a policy was
the average score reached in three consecutive games. Selection
ratio and step size were set to $\rho = 0.05$ and $\alpha = 0.6$,
respectively. Furthermore, in each iteration during learning, we
slowly decayed the slot usage probabilities $p_i$ with decay
factor $\beta = 0.98$. This choice slightly biased the
optimization towards shorter policies.

\subsection{Automatically constructed rulebase}

In this experiment, we applied the same policy selection mechanism
as in the previous experiment, but we did not use a hand-coded
rulebase. At the beginning of learning, rules were drawn randomly
for each $(i,j)$ pair with $i\in \{1,\ldots,m\}$ and $j\in
\{1,\ldots,K\}$. A random rule is a random pair of a randomly
drawn condition set and a randomly drawn action. Random condition
sets contained 1, 2, or 3 conditions. These rules were not changed
during learning, only their corresponding probabilities were
optimized.

The following parameter values were used: population size:
$N=1000$, number of rule slots: $m=90$, number of possible rules
in each slot: $K=100$, selection ratio: $\rho = 0.05$, step-size:
$\alpha = 0.6$, decay rate: $\beta=0.98$.

\subsection{Baseline experiments}

A large amount of domain knowledge was used while constructing the
high-level observations and actions, which is obviously a key
factor in reaching good performance. In order to isolate and
assess the contribution of learning, we performed two additional
experiments with different amounts of domain knowledge and no
learning.

In the first non-learning experiment, we used the rulebase of 40
hand-coded rules (identical to the rulebase of the first learning
experiment). Ten rules were selected at random, and random
priorities were assigned to them. We measured the performance of
policies constructed in this way.

In the second non-learning experiment, we hand-coded a full policy
(both rules and priorities). The policy is shown in
Table~\ref{t:policyexample}, and has been constructed by some
trial-and-error. Naturally, the policy was constructed before
knowing the results of the learning experiments.

In the final experiment, five human subjects were asked to play
the first level of Pac-Man and we measured their performances.
Each of the subjects has played Pac-Man and/or similar games
before, but none of them was an experienced player.

\section{Experimental results} \label{s:results}

Human experiments were performed on the first level of an
open-source Pac-Man clone of \citet{Courtillat01Pacman}. For the
other experiments we applied the Delphi re-implementation of the
code.

In both learning experiments, 10 parallel learning runs were
executed, each one for 300 episodes. This training period was
sufficient to tune all probabilities either to 0 or 1, so the
learned policy could be determined in all cases. Each obtained
policy was tested by using it for 50 consecutive games, giving a
total of 500 test games per experiment.

In the non-learning experiments the agents played 500 test games,
too, using random policies and the hand-coded policy,
respectively. Each human subject played 20 games, giving a total
of 100 test games.

\begin{table}
\begin{center}
\caption{Pacman results. Maximum available score is 13900. See
text for details.}
 \label{t:results} \vskip 0.15in
\begin{tabular}{l|c|c|c}
   Method  & Mean & High & \# of rules \\[2pt]
\hline \hline \rule{0pt}{12pt}%
  Random rulebase + CE & 6312 & 13900 & 3.9 \\
  Hand-coded rulebase + CE & 7636 & 13900 & 8.0 \\ \hline
  Hand-coded rulebase + random rules & 257 & 2010 & 10 \\
  Hand-coded policy & 5670 & 10660 & 9 \\
  Human play & 8064 & >13900\footnote{f} & - \\
 \hline
\end{tabular}
\end{center}
\end{table}

\footnotetext{Humans could occasionally score 100 points by
`eating' fruits. This option was not implemented in the
machine-play version.}

Both the average scores and the high scores are summarized in
Table~\ref{t:results}. Comparing scores for hand-coded domain
knowledge with and without learning, we found that the
contribution of cross-entropy learning is significant. The average
number of rules in the learned policies shown in the last column
of the table, varies. Policies found by the learning methods
performed better than the hand-coded policies \emph{and} they were
shorter on the average.

On the other hand, the learned policies are still far from being
optimal, and could not reach the level of non-experienced human
players. We investigated how the game is played by various
policies in order to identify the possible reasons of superior
human performance. It seems that the major flaw of the rule-based
policies is that they cannot eat all ghosts when the ghosts turn
blue. This is a serious handicap. For example, if the agent can
eat only three ghosts after ghosts turn blue, but otherwise plays
perfectly, it can only reach $13900 - 4\cdot1600 = 7500$ points.
The task of catching all ghosts in a limited time period can be
successful only if all the ghosts are nearby, and this requires
strategic planning: power dots should be eaten only after all
ghosts have been lured close to it. The set of available high
level observations does not enable such planning: the agent cannot
observe how scattered the ghosts are or how far the farthest ghost
is. This type of information is easily available for human
players, who `see' the board and observe the topological structure
of the maze.

\section{Discussion} \label{s:discussion}

\subsection{The role of domain knowledge}

When demonstrating the abilities of an RL algorithm, it is often
required that learning starts from scratch, so that the
contribution of learning is clearly measurable. However, the
choice of test problem is often misleading: many `abstract'
domains contain considerable amount of domain knowledge in an
implicit way. As an example, consider gridworld navigation tasks,
an often used class of problems for `tabula rasa' learning. In a
simple version of the gridworld navigation task, the state is an
integer that uniquely identifies the position of the agent, and
the atomic actions are moves to grid cells north/south/east/west
from the actual cell.

The concepts of north, south, etc. corresponds to very high-level
abstraction, they have has a meaning to humans only, so they are
domain knowledge. In fact, they are very similar to the domain
knowledge provided by us, the high-level observations and actions:
observations like `distance of nearest ghost is $d$' or 'Pac-Man
is at position $(11,2)$' are both high-level observations.
Similarly, action 'go north' and action 'go towards the nearest
power dot' are essentially of the same level.

The implicit presence of high-level concepts becomes even more
apparent as we move from abstract MDPs to the `real-world'.
Consider a robotic implementation of the maze task: the
observation of the state is not available for the robot. It sees
only local features and it may not see all local features at a
time. To obtain the exact position, or to move by one unit length
in the prescribed direction, the robot has to integrate
information from movement sensors, optical/radar sensors etc. Such
information fusion, although necessary, but does not belong to
reinforcement learning. Thus, in this task, there is a great
amount of domain knowledge that needs to be provided before our CE
based policy search method could be applied.


Naturally, assessing the effectiveness of a learning algorithm is
more difficult for non-abstract tasks, because we have to measure
the contribution of human knowledge somehow. Our experiments with
random and hand-picked policies intend to estimate the
contribution of (a varying amount of) human knowledge.

In our opinion, the role of human knowledge is that it selects the
set of observations and actions that suit the learning algorithm.
Such extra knowledge is typically necessary for most applications.
Nonetheless, numerous (more-or-less successful) approaches exist
for obtaining such domain knowledge automatically. According to
one approach, the set of observations is chosen from a rich (and
redundant) set of observations by some feature selection method.
The cross-entropy method seems promising here, too \citep[see][
for an application to feature selection from brain fMRI data at
the 2006 Pittsburgh Brain Activity Interpretation
Competition]{Szita06How}. According to a different approach,
successful combinations of lower level rules can be joined into
higher level concepts/rules. Machine learning has powerful tools
here, e.g., arithmetic coding for data compression
\citep{witten87arithmetic}. It is applied in many areas, including
the writing tool Dasher developed by \cite{ward02fast}. Such
extensions are to be included into the framework of reinforcement
learning.

\subsection{Low-complexity policies}

The space of legal policies is huge (potentially infinite), so it
is an interesting question how search can be effective in this
huge space. Direct search is formidable. We think that an implicit
bias towards low-complexity policies can be useful. Solutions can
be used as building blocks in a continued search of low-complexity
policies. Low-complexity policy here means that even if a policy
consists of very many rules, in most cases, only a few of them is
applied in the game.\footnote{Of course, it is possible to
construct long policies so that each rule gets applied. However,
the chance is tiny that we find long policies by random sampling.}
Unused rules do not get rewarded (nor do they get punished unless
they limit a useful rule), so the \emph{effective length} of
policies is biased towards short policies. This implicit bias is
strengthened by an explicit one in our work: the probabilities of
application of a rule decay, so indifferent rules get wiped out
soon.

The bias towards short policies reduces the effective search space
considerably. Further, for many real-life problems, low-complexity
solutions exist \citep[for an excellent analysis of possible
reasons, see][]{Schmidhuber97Computer}. Therefore, search is
concentrated on a relevant part of the policy space, and pays less
attention to more complex (and therefore less likely) policies.

\subsection{Summary and Outlook}

In this article we proposed a method that learns to play Pac-Man.
We have defined a set of high-level observation and action modules
with the following properties: (i) actions are temporally
extended, (ii) actions are not exclusive; actions may work
concurrently. Our method can uncover action combinations together
with their priorities. Thus, our agent can pursue multiple goals
in parallel.

The decision of the agent concerns whether an action module should
be turned on (if it is off) or off (if it is on). Further,
decisions depend on the current observations and may depend on the
state of action modules. The policy of the agent is represented as
a list of if-then rules with priorities. Such policies are easy to
interpret and analyze. It is also easy to incorporate additional
human knowledge. The cross-entropy method is used for learning
policies that play well. Learning is biased towards low-complexity
policies, which is a consequence of both the policy representation
and the applied learning method. The learned policies reached
better score than the hand-coded policy, and neared the score of
average human players.

The applied architecture has the potentials to handle large,
structured observation- and action-spaces, partial observability,
temporally extended and concurrent actions. Despite its
versatility, policy search can be effective, because it is biased
towards low-complexity policies. These properties are attractive
from the point of view of large-scale applications.

\subsubsection*{Acknowledgments} This material
is based upon work supported partially by the European Office of
Aerospace Research and Development, Air Force Office of Scientific
Research, Air Force Research Laboratory, under Contract No.
FA-073029. This research has also been supported by an EC FET
grant, the `New Ties project' under contract 003752. Any opinions,
findings and conclusions or recommendations expressed in this
material are those of the author(s) and do not necessarily reflect
the views of the European Office of Aerospace Research and
Development, Air Force Office of Scientific Research, Air Force
Research Laboratory, the EC, or other members of the EC New Ties
project.

\newpage

\part*{Appendix:  The low-complexity Modular Policy Framework}

\section{A critique of Markov decision processes}

Modelling RL problems as (finite) Markov decision processes (MDPs)
has proved very fruitful both in the theoretical grounding and in
some practical applications. However, because of the
simplifications of the MDP model, such as full observability, the
Markov property, finite and unstructured state- and action space,
equal sized time steps etc., it does not scale well for typical
``real-life" applications. Therefore, most of the recent research
in RL tries to extend the MDP framework in various directions or
tries to find alternative models.

The MDP model is too general in some respects as it has been noted
e.g. in \cite{lane05why}: an RL algorithm is expected to solve
\emph{any} MDPs (at least approximately) in the same manner, and
it is well known that this cannot be done faster than polynomial
in the number of the states. However, practical problems often
have billions of states and polynomial time solutions are
intractable. Nonetheless, many of these problems have compact
structured descriptions that might enable more specific
algorithms. We also note that computational intractability, e.g.,
the ``curse of dimensionality", severely restricts MDPs and its
extensions, e.g., partially observable MDPs (POMDPs), predictive
state representations, observable operator models, semi-MDPs, with
a few notable exceptions like factored MDPs.

We collect here several requirements that have to be resolved for
large-scale, ``real-life" RL tasks. We argue that these
requirements can be handled in a unified way, provided that the
attributes of the agent, such as action, state, and memory, are
treated on equal footings, and that the agent is characterized by
a (factored) set of modules. Each of these modules may be
state-like, action-like etc., or even the mixture of these. We
show that in this formalism, policy is a module to module mapping
that makes mathematics simple. This is true even for complex
policies involving partial observability, memory management,
attention focusing or parallel actions, issues that emerge in many
practical problems.

We also show that if the complexity of the policy is low then, in
our formalism, the learning task becomes tractable \emph{without}
further compromises. We provide an algorithm that learns
low-complexity modular policies in the form of decision queues,
show that it is convergent, and -- in the idealistic limit case --
it finds the optimum.

\section{Modular representation: An informal description}

\subsection{An illustrative example}

Let us consider driving a car in the city in order to list the
challenges of real-life RL agents. When driving towards a
destination, the driver has to cross intersections, has to pass
other cars, and has to obey the traffic signs and traffic lights.
Unfortunately, the driver cannot observe everything about its
current situation, e.g. if one looks to the left, she cannot see
what is on the right; if she looks at the mileometer, she cannot
see what is happening on the road (\emph{partial observability}).
She decides where to look depending on the situation: at the car
before her, the traffic signs, the control panel, or something
else (\emph{attention focusing}). She is aided by her short-term
\emph{memory}: she remembers recent observations. She is engaged
in \emph{parallel multiple activities}: steers and speeds up for
an overtake, uses the brake and looks around in a crossing. Such
combined actions are typical in driving. The durations of the
actions and events may vary, and are not well defined
(\emph{non-uniform time steps}). Also, actions can be continuous,
like braking, or discrete, like switching the lights on.
Similarly, observations can also be continuous, like the distance
from the crossing, or discrete, like the color of traffic light.

Although the policy of the driver is very complex, only a tiny
fraction of the possible policies is ever tried. For example, most
drivers may never try to find out the immediate reward for looking
right, pushing the brake, steering the wheels to the right,
getting into a small street, then getting the car straight, to
speed up and look back, not to mention other combinations, like
looking right and turning the wheels left. Despite the complexity
of the policy, it is built up from simple ones by means of simple
combination rules.

\subsection{Modules}

As the above description illustrates, all the attributes of the
agent, its observations, actions and memory have much in common:
(1) they have factorized structure (2) they can be either
continuous or discrete (3) they can be influenced by the policy
and (4) the policy can be influenced by them. Furthermore, the
distinction between them is blurred, e.g. the action `turn back to
see what's behind me' is an observation, manipulates memory, and
focuses attention (what to observe). Therefore it seems reasonable
to treat them as different forms of a single concept that we will
call \emph{modules}. We can talk about observation-like,
memory-like or action-like modules, but these concepts are not
necessarily exclusive.

Using such a representation, the agent is described by a set of
modules. These modules constitute a factored representation, and
their domain is arbitrary (continuous or discrete). Naturally, the
agent itself can modify only some of its modules, others are
influenced by its environment and again, these two sets are not
exclusive.

\subsubsection{Preserving computational tractability}

Because of the factored structure, it is possible that the
decisions of the agent depend only on a few modules, and affect
only a few other ones. Therefore, we have the opportunity to
express simple activities with simple (short) policy descriptions.
This enables us to make the policy search tractable by restricting
search to simple policies.

We shall define 'simplicity' rigorously in the next section.
Basically, we are looking for policies that are composed of
relatively few decisions and these decisions have compact
descriptions about their conditions and their effects. Then we can
manage the search, which is polynomial in the size of the problem
description. This restricted policy space still contains
interesting policies: many real-life solutions have simple
structures, despite of the size of the state space. Problems with
complex \mbox{(near-)}optimal solutions are hard for humans, too,
and they are outside of our present considerations.

\subsection{Advantages of modular representation}

Below we summarize the expected advantages of the modular
representation. Firstly, we are able to handle partial
observability, memory, and in particular, focus of attention.
Secondly, because of the unified treatment of various agent
attributes, policies assume a simple form despite of their
complexities compared to, e.g. memoryless MDP policies.
Furthermore, we can handle composite attributes, e.g. a single
module may have observation-like and action-like components. The
factored representation enables us to use multiple state variables
and/or multiple parallel actions and, in turn, many interesting
problems may have compact descriptions. And finally, we can use
differential policy representation (the policy prescribes how to
\emph{change} the actual representation), simple policies can have
compact descriptions. Thus, we can restrict our searches to the
set of simple policies.

\subsection{Related literature}

Due to the limitations of space, we can mention only the most
relevant frameworks and methods.

The general framework for handling partial observability is the
POMDP framework \cite{murphy00survey}, but recently, other
alternatives were also proposed, including predictive state
representations \cite{singh03learning} and observable operator
models \cite{jaeger99action}. In POMDPs, memory and attention are
handled implicitly, but there are also numerous methods that use
explicit memory management, e.g. memory bits, finite state
machines, variable-length history suffixes, or attention focusing.
Another direction that extends MDP is the semi-MDP (SMDP)
framework \cite{sutton99between}, which enables e.g. the use of
parallel, varying-length actions (although they must be
synchronized). SMDP is also used in hierarchical methods
\cite{barto03recent}.

These models are all extensions of MDPs, so general solution
algorithms for them are computationally at least as intractable as
for MDPs or may be even harder. Function approximation (FAPP) and
direct policy search (see e.g. \cite{sutton00policy}) are two
common and successful techniques for reducing complexity. However,
policies learnt by policy search and/or FAPP keep many of the MDP
restrictions; they are memoryless, use reactive policies, and can
not handle parallel and varying-length actions. Furthermore,
constraints of the parameter space introduce other restrictions
that are often non-intuitive.

In our approach, state space representation is similar to factored
MDPs (e.g. \cite{guestrin03efficient}), but we proceed by policy
search instead of learning value functions.

\section{Formal description of the low-complexity modular policy framework}

Often, non-modifiable components, such as the value of an
observation, or the execution of a longer action, have related
components that control its usage, e.g. if we can observe that
variable, if the action is running or not, or if the relevance of
the component is high or low for the agent. Therefore, it seems
practical to define modules as pairs, consisting of (i) the output
value of the module, and (ii) the extent that the module is used
or whether it is used at all. In principle, the range of output
values can be from an arbitrary set, but for the sake of
simplicity, we restrict it to (subsets of) real numbers. Also, we
can restrict modules to on and off states $\{0,1\}$ that can be
switched, or we can use real numbers to represent their influence,
which can be tuned on a continuous scale.


\begin{defn}[Module] a pair $(w, x)$ is called a module, where $x\in
\Real$ is the actual output value of the module, and $w \in \Real$
is its influence..
\end{defn}

\begin{defn}[Modular state representation]
For $m\ge1$, the ordered set $((w_1,x_1),$ $\ldots,$ $(w_m, x_m))$
is called an $m$-dimensional modular state representation, if
$(w_i, x_i)$ is a module for any $1\le i \le m$. The set of
modular state representations for a fixed $m$ is denoted by
$\mathcal M$.
\end{defn}

Let $\Pi$ be the set of all $\mathcal M \to \mathcal M$ mappings.
A modular policy $\pi$ that belongs to $\Pi$ can be subject to
restrictions. For example, we may ensure that the policy is
constrained to legal actions and the agent does not execute
actions that are unsafe or contradictory, it cannot modify the
actual values of the observations etc. Such constraints will be
encoded by a problem-specific mapping $\delta: \mathcal P \times
\mathcal P \to \mathcal P$. $\delta$ is the \emph{internal
dynamics} and maps the current representation and the one proposed
by the policy to the realized state representation. We shall also
limit the complexity of the policies; subset $\Pi_0$ will denote
the set of `simple' policies (see later).

We define the environment of the agent as a general controllable
dynamic process that provides observable quantities and rewards.
We do not assume anything, e.g. full observability, beyond that.

\begin{defn}[Environment]
\setlength{\itemsep}{-2pt}

Let $S$, $O$ and $A$ be arbitrary state, observation and action
spaces, respectively. The environment is a tuple $(s_0, \sigma,
\omega, \rho)$, where
\begin{itemize}
\setlength{\itemsep}{-2pt} \item $s_0 \in S$ is the initial state,

\item $\sigma: S \times A \times S \to [0,1]$ is the transition
function of the environment,

\item $\omega: S \times O \to [0,1]$ is the observation function,

\item $\rho: S \to \Real$ is the reward function.
\end{itemize}
\end{defn}

The agent is determined by its policy, its internal dynamics, and
the interfaces that map primitive observations to modules and
modules to primitive actions (and may handle conflicting actions).

\begin{defn}[Modular representation agent]
For a given observation space $O$ and action space $A$, a modular
representation agent is a tuple $(\b m_0, \phi, \delta, \psi,
\pi)$, where
\begin{itemize}
\setlength{\itemsep}{-2pt} \item $\b m_0 \in \mathcal M$ is the
initial module representation,

\item $\phi: O\times \mathcal M \to \mathcal M$ is the input
interface that tells the effect of observations on the modules of
the agent,

\item $\psi: \mathcal M \to A$ is the output interface that
translates modules to primitive actions,

\item $\pi: \mathcal M \to \mathcal M$ is the policy of the agent,

\item $\delta: \mathcal M \times \mathcal M \to \mathcal M$ is the
internal dynamics of the agent.
\end{itemize}
\end{defn}

With these definitions, we can formally describe the
agent-environment interaction: consider an environment $E = (s_0,
\sigma, \omega, \rho)$ and a modular representation agent $G = (\b
m_0, \phi, \delta, \psi, \pi)$. At $t=0$, the environment is in
state $s_0$ and the agent is in state $\b m_0$. The interaction is
as follows:
\begin{eqnarray*}
  o_t \sim \omega(s_t,.)  && \textrm{(observation)} \\
  r_t := \rho(s_t)  && \textrm{(reward)} \\
  \b m'_t := \phi(o_t,\b m_t)    && \textrm{(observation-and-module-to-module mapping)} \\
  \Delta\b m_t := \pi(\b m'_t)    && \textrm{(decision of the agent)} \\
  \b m_{t+1} := \delta(\b m'_t, \Delta \b m_t)    && \textrm{(internal dynamics)} \\
  a_{t+1} := \psi(\b m_{t+1})    && \textrm{(module-to-action mapping)} \\
  s_{t+1} \sim \sigma(s_t,a_{t+1},.)    && \textrm{(environment dynamics)} \\
\end{eqnarray*}

The decision task can be formalized by fixing the parameters of
the environment and the interface:

\begin{defn} A modular sequential decision problem is given by
an environment $E$, a set of allowed policies $\Pi_0$ and a family
of agents $\{G(\pi) : \pi \in \Pi_0\}$ with fixed interface
mappings and internal dynamics, and a discount factor $0\le \gamma
\le 1$.

A solution of this problem is a policy $\pi^*$ for which the
expected discounted cumulative reward,
\[
  E(r_0 + \gamma r_1 + \gamma^2 r_2 + \ldots)
\]
is maximal, supposed that the system is working according to the
equations above.
\end{defn}

\subsection{Low-complexity modular policies}

We have to define a restricted policy set $\Pi_0$. There are
different approaches for bounding complexity: one can describe
policies with a fixed (small) number of parameters (used e.g. in
policy search methods), decision trees, or decision queues. As an
example, we shall apply decision queues here, which is a flexible
structure and fits nicely into the general optimization algorithm
to be utilized.

A decision queue is an unordered list of rules, where every rule
assumes the form
\[
  \verb"[priority]": \quad \textbf{if } \textit{Cond}(\b m_t) \textbf{ then }
  \Delta \b m_t := \pi^a(\b m_t),
\]
where $\textit{Cond}(\b m_t)$ is a Boolean expression depending on
the current module representation, $\pi^a$ is a policy, which is
considered \emph{atomic}, and $\verb"priority" \in \{1,\ldots,K\}$
determines the order of the rules in the queue. The action taken
by a decision queue is determined by checking all the rules in the
order of their priorities (ties are broken arbitrarily). We choose
the first rule with satisfied conditions, and execute its
prescribed atomic policy.

To achieve low complexity, both the conditions and the atomic
policies are chosen from a finite set with polynomial size in the
number of modules, and the number of priorities is also kept low.
This ensures that the building blocks have simple (short)
encodings. Furthermore, the number of building blocks in a queue
will be also limited. Policies of this kind will be called
\emph{low-complexity modular policies} (LCM policies).

\section{Finding optimal LCM policies}

Let $R$ be the set of possible rules and $N$ be the maximum number
of allowed rules in a policy. For all $n\in [1,\ldots,N]$, let
$R_n \subseteq R$ be a subset of applicable rules belonging to
index $n$ and \mbox{$P_n \subseteq \{1,\ldots,K\}$} is the subset
of applicable priorities. Let $\Pi_0$ be the set of allowed
priority queues. To apply the CE method, we define a distribution
over $\Pi_0$: in episode $z$, let us denote the probability that
rule $n$ will be selected by $p^{(z)}_n$. If rule $n$ is used, we
have $|R_n|$ choices of rules to choose from. The probability of
the $i^{th}$ one is denoted by $q^{(z)}_{ni}$. We draw from $2N$
independent Bernoulli distributions with $2, \ldots, 2, |R_1|,
\ldots, |R_N|$ choices. We can directly apply the CE method to
them: in each episode, we draw a population of policies according
to the current distribution, try them to get their cumulated
reward, select the elite, and use Eq.~\ref{e:CEupdate_bernoulli}
to update the distribution.

We prefer short policies: probabilities $p^{(z)}_n$ are discounted
by a factor $\beta<1$ in each step.

\subsection{Convergence}

It is known that under mild regularity conditions, the CE method
converges with probability 1 \cite{margolin04onthe}. Furthermore,
for a sufficiently large population, the global optimum is found
with high probability.

The CE method has become attractive through a large number of
experimental evidences that -- even with small populations -- it
finds good local optima of large, hard instances of NP-hard
problems (cf. references in \cite{Boer04Tutorial}). Also,
performance is insensitive to the particular choice of
optimization parameters in a broad range, so little fine-tuning is
necessary.

\section{Applying the LCMP Framework to Pac-Man}

As we could see, some kind of processing of the input (and
possibly output) is necessary to make the learning problem
tractable. This is easy; one can implement potentially useful
features and primitive actions similar those applied by human
players, e.g. `the distance of the nearest ghost/pellet/power
pellet', `average distance of ghosts/pellets', `length of current
corridor', `go towards the nearest pellet/power pellet/edible
ghost', `keep direction', `go away from nearest ghost', etc. The
real challenge is how to utilize these modules and how to combine
them.

These features are inherently continuous and modular, some of the
actions may run side-by-side (e.g. `go to nearest pellet' and
`keep direction'), others may conflict, and their duration may
vary. Any of the observations may prove useful in certain
situations, but the agent will never need all of them at once. All
of these properties are in concordance with the LCMP framework and
this framework can be readily applied if features and primitive
actions are all treated as modules. Note that Pac-Man's policy may
be non-Markovian, because CE does not exploit the Markov property.

\section{Discussion}

The LCMP framework provides a  general model for formalizing
reinforcement learning problems. In this model, the agent's state
representation is a set of parallel modules that can be switched.
Modules unify observations, actions and memory in a mathematically
simple, general concept. We showed that modular policies satisfy a
number of desirable requirements in a natural way. By bounding the
complexity of modular policies, the learning problem becomes
tractable. To demonstrate this, we described an applicatoin of the
framework to Pac-Man.

We note that our formalism allows one to provide a large amount of
pre-wired knowledge (such as those used in the Pac-Man
experiments). For many real-life problems, such knowledge is
easily available, and we believe that it is also necessary for
obtaining good performance. The problem, how to emerge high-level
concepts by machine learning is out of the scope of the present
study. We also note that we are not aware of any RL method that
would be able to handle the large state space, partial
observability, parallel and varying-length activities that are
present in the full-scale Pac-Man game.

Exploration of the potentials of LCM policies is still at an early
stage, so there are many open questions. For example, it is
unclear how to perform credit assignment, i.e. how to decide the
contribution of a given rule to the total performance of the
policy. Bucket brigade-like methods applied in evolutionary
methods \cite{bull05foundations} seem promising here.

\bibliographystyle{plainnat}
\bibliography{psmdp}

\end{document}